\documentclass[a4paper,twoside]{article}

\usepackage{epsfig}
\usepackage{subfig}
\usepackage{calc}
\usepackage{amssymb}
\usepackage{amstext}
\usepackage{amsmath}
\usepackage{amsthm}
\usepackage{multicol}
\usepackage{pslatex}
\usepackage{apalike}
\usepackage{glossaries}
\usepackage{SCITEPRESS}

\newacronym{fov}{FoV}{Field-of-View}
\newacronym{soc}{SoC}{System-on-Chip}
\newacronym{fps}{FPS}{Frames per second}
\newacronym{nms}{NMS}{non-maximum suppression}
\newacronym{iou}{IoU}{Intersection over Union}
\newacronym{flops}{FLOPS}{floating point operations}

\begin{document}

\title{How low can you go?\subtitle{Privacy-preserving people detection with an omni-directional camera.}}

\author{\authorname{Timothy Callemein, Kristof Van Beeck, and Toon Goedem\'{e}}
\affiliation{EAVISE, KU Leuven, Jan Pieter de Nayerlaan 5,Sint-Katelijne-Waver, Belgium}
\email{ \{timothy.callemein, kristof.vanbeeck, toon.goedeme\}@kuleuven.be}}


\abstract{In this work, we use a ceiling-mounted omni-directional camera to detect people in a room. This can be used as a sensor to measure the occupancy of meeting rooms and count the amount of flex-desk working spaces available. If these devices can be integrated in an embedded low-power sensor, it would form an ideal extension of automated room reservation systems in office environments.
The main challenge we target here is ensuring the privacy of the people filmed. The approach we propose is going to extremely low image resolutions, such that it is impossible to recognise people or read potentially confidential documents. 
Therefore, we retrained a single-shot low-resolution person detection network with automatically generated ground truth. In this paper, we prove the functionality of this approach and explore how low we can go in resolution, to determine the optimal trade-off between recognition accuracy and privacy preservation.
Because of the low resolution, the result is a lightweight network that can potentially be deployed on embedded hardware. Such embedded implementation enables the development of a decentralised smart camera which only outputs the required meta-data (i.e. the number of persons in the meeting room).
}

\onecolumn \maketitle \normalsize \vfill

\begin{figure*}
    \centering
    \includegraphics[width=\linewidth]{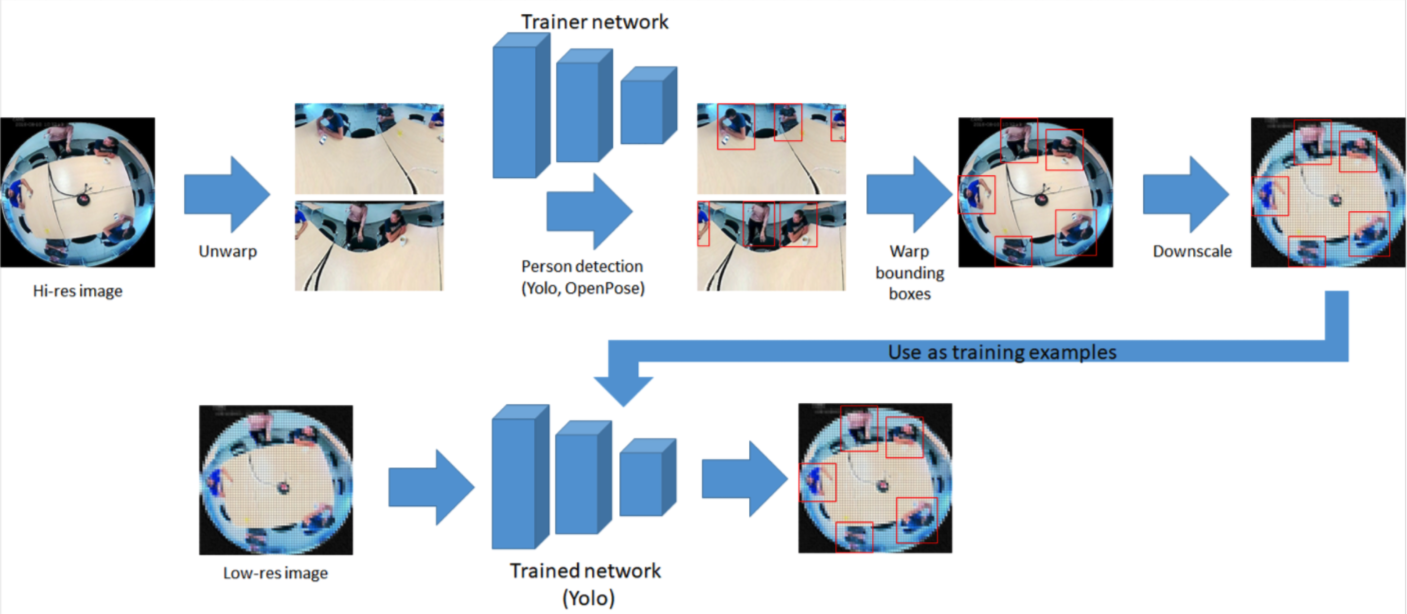}
    \caption{The overall system of the automated annotation process and the retraining of the YOLOv2 models }
    \label{fig:overall_system}
\end{figure*}

\section{Introduction}
\label{sec:introduction}

Recent progress in deep learning has provided researchers with many exciting possibilities that pave the way towards numerous new applications and challenges.
Apart from the academic world, the industry has also noticed this evolution and shows interest in adopting these techniques as soon as possible in their real-life cases.
The field of computer vision is a fast-changing landscape, constantly improving in speed and/or accuracy on publicly available datasets.
However, industrial companies frequently construct their own datasets which are often not -- or incorrectly -- annotated and unavailable for public use.
This paper aims to tackle a very specific real-life scheduling problem: automatically detecting the occupancy of meeting rooms or flex-desks\footnote{A space where you do not have a fixed desk but use the available desks, commonly with a reservation list.} in office buildings, using only RGB cameras that are mounted on the ceiling.
Indeed, in practice it is often the case that the occupancy of the available meeting rooms is sub-optimal (i.e. a large meeting room is occupied with only few participants).
On a large scale, this has a significant economic impact (e.g. the construction of additional office buildings could be avoided if the occupancy is improved).
Such problems are mainly due to employees having a reoccurring reservation on a flex-desk while working from home or empty booked meeting rooms when the meeting got cancelled.
Through the occupancy detection of each meeting room (and flex-desk) companies aim to optimise their office space capacity.
Not only for binary cases, used and not used, but also to optimise the capacities of the meeting rooms by counting the amount of people during a meeting, opposed to the meeting room capacity.

However, placing cameras in working environments or public places inherently involves privacy issues. Such issues are strictly regulated by the governmental bodies which in most countries allow such recordings when they are not transmitted to a centralised system or saved in long term memory.
It is however allowed to derive, save and transfer generated meta-data based on images containing people.
Apart from these regulations, the employees present will have a feeling of unease when a camera is watching their movements, even when the company claims only meta-data is being transmitted.
To cope with the aforementioned problem, we propose to reduce the resolution of the input images (or even the physical number of pixels on the camera sensor itself) such that persons become inherently unrecognisable.
Evidently, there is a lower limit in resolution down-sampling: at a specific point the computer vision algorithms fail to efficiently detect persons. As such, we need to determine the optimal resolution at which persons are unrecognisable (to human observers) yet are still automatically detectable.

One of the main goals of this paper is to evaluate \textit{how low we can go} in resolution until the deep learning algorithms fail to detect people. We thus aim to investigate the trade-off between input image resolution (i.e. privacy), detection accuracy and detection speed.
By lowering the amount of input data the detection challenge for the neural network is increased, but a gain in speedup is achieved since the amount of required calculations decreases dramatically.
Indeed, deep learning architectures often require significant computational power to perform inference.
A lower computational complexity allows for an embedded implementation on the camera itself (i.e. a decentralised approach) where the original image data never leaves the embedded device. Such scenario would be ideal as this inherently preserves privacy.

In this paper -- as a proof-of-concept -- we start from a standard camera with high resolution and manually downscale the images in order to compare performances in-between different resolutions.
During deployment however, the system will first learn in high-resolution and can be replaced by a low-resolution for inference only.
For example placing the hardware lens out of focus, resembles a blur filter and is considered a hardware adjustment.
An additional challenge arises from the fact that as few cameras as possible should be required to cover the meeting rooms.
Therefore, we use omni-directional cameras recording the 360$^\circ$ space around the camera.
For this, we rectify the image, as discussed further on.
Finally, the acquisition of enough annotated training data to (re)train deep neural networks remains time-consuming and thus expensive.
In this paper we propose an approach in which we automatically generate annotations to retrain our low-resolution networks, based on the high resolution images. Figure \ref{fig:overall_system} shown an overview of this approach.
To summarise, the main contributions of this paper are:
\begin{itemize}
    \item Finding an optimal trade-off between: lowest resolution, accuracy, processing time, perseverance of privacy
    \item An automatic pipeline where we run two detectors on unwarped images to output annotations
    \item Public Omni-directional dataset containing several meeting room scenarios
\end{itemize}
The remainder of this paper is structured as follows. We first talk about the related work in section \ref{sec:related_work} followed by discussing what we recorded and used as dataset in section \ref{sec:dataset}.
In section \ref{sec:automatic_annotations} we detail the top row in fig.~\ref{fig:overall_system}: the automatic generation of automatic annotations on our dataset, where we use a combination of strong person detectors on the high-resolution images to find reliably the persons in the room. The latter step yields annotated images, which we downscale in resolution to use as training examples for our low-resolution person detector, illustrated in the bottom row of fig.~\ref{fig:overall_system} and explained in further detail in section \ref{sec:howlowcanwego}.  showing our results on how low we can go using the annotated data. 
Section \ref{sec:conclusion} will conclude this research and show some possible future work based on our current results.

\section{Related Work}
\label{sec:related_work}

Employing cameras in public or work environments inherently presents privacy issues.
People tend to feel uneasy given the knowledge that they are constantly being filmed.
Furthermore often sensitive documents are processed in the work environment.
Several possible solutions exist which aim to temper these problems. First, a closed system could be constructed in which the sensitive image data never leaves the device.
However, even if this is the case most people remain reluctant to the use of these devices.
A second solution could be the use of an image sensor with an extremely low resolution such that privacy is automatically retained.
Such solution could be mimicked using a high resolution camera of which the images are e.g. blurred or down-sampled.
A user study presented in \cite{butler2015privacy} shows that the use of different image filters indeed increases the sense of privacy.
Even a simple blur image filter of only $5px$ already decreased the privacy issues (depending on the object which was visible).
In this work we downsample the image which gives similar results as a blur filter (concerning privacy).
We aim to evaluate several resolutions and expect the sense of privacy to increase with every downsampling step.

In section~\ref{sec:dataset} we discuss how we recorded a new dataset which will be made publicly available meeting all of our criteria.
A large disadvantage when recording a new dataset is found in the manual labour to annotate all image frames.
Instead of resorting to time-consuming manual annotations we propose the use of knowledge distillation to automatically generate the required annotations.
We thus aim at transferring knowledge from a teacher network to a student network. 
In fig.~\ref{fig:overall_system} the top row will represent the teacher architecture, while the bottom row is the student network that will use the generated annotations by the teacher as training data.
Techniques presented in \cite{hinton2015distilling,ba2014deep,lee2018self} indeed show the potential of these approaches and illustrate that it is possible to use a pretrained complex network to train a simple network.
Our input data consists of omni-directional images and thus no pretrained person detector models are currently available.
Therefore we cannot use knowledge distillation in its current form.
Instead of using a complex network to re-purpose the weights of a new model, we aim to employ person detectors on the unwarped high-resolution omni-directional images. 
These detections are then used as training examples for the low-resolution networks.
Several excellent state-of-the-art object detector exist. 
For example, SSD \cite{liu2016ssd}, RetinaNet \cite{he2016deep}, R-FCN \cite{dai2016r} all perform well with high accuracy and are capable of running real-time on modern desktop GPUs.
In this paper we opted for the Darknet framework \cite{redmon2017yolo9000} (more specifically, the YOLOv2 architecture). 
This framework outperforms the aforementioned methodologies in terms of speed, by a factor 3 or more with a negligible loss in accuracy.
However, our initial experiments revealed that in specific cases the person detection fails, mainly at regions with heavy lens distortion.
Therefore we used a single-frame bottom-up pose estimator \cite{cao2017realtime,wei2016cpm} to further increase the accuracy and efficiently fused both methodologies as discussed further.
While the person detector focuses on the overall person, the pose estimator first detects separate body parts which are then combined into a complete pose.

In a next step the combined person detector approach mentioned above is used to automatically generate annotations on the high resolution input images.
These annotations are then used to train the low-resolution networks.
Similar work is proposed in \cite{chen2017semi} where the authors uses high resolution frames combined with extreme low resolution frames to train a single model.
The showed that the combination of both feature spaces produce an action recognition model with few parameters.
Different work by \cite{ryoo2017privacy} also focuses on extreme low resolution images for action recognition, but uses video as an additional dimension.
Both works show that even at extreme low resolutions ($12\times16$) action recognition based on down-scaled frames is possible.
Our work significantly diverges from previous works.
We aim to develop a frame-by-frame system to detect people  and count people (instead of preforming action recognition) with an emphasis on finding an optimum trade-off between a privacy preserving low resolution and high accuracy.

\section{Dataset}
\label{sec:dataset}

\begin{figure*}[ht]
    \centering
    \includegraphics[width=\linewidth]{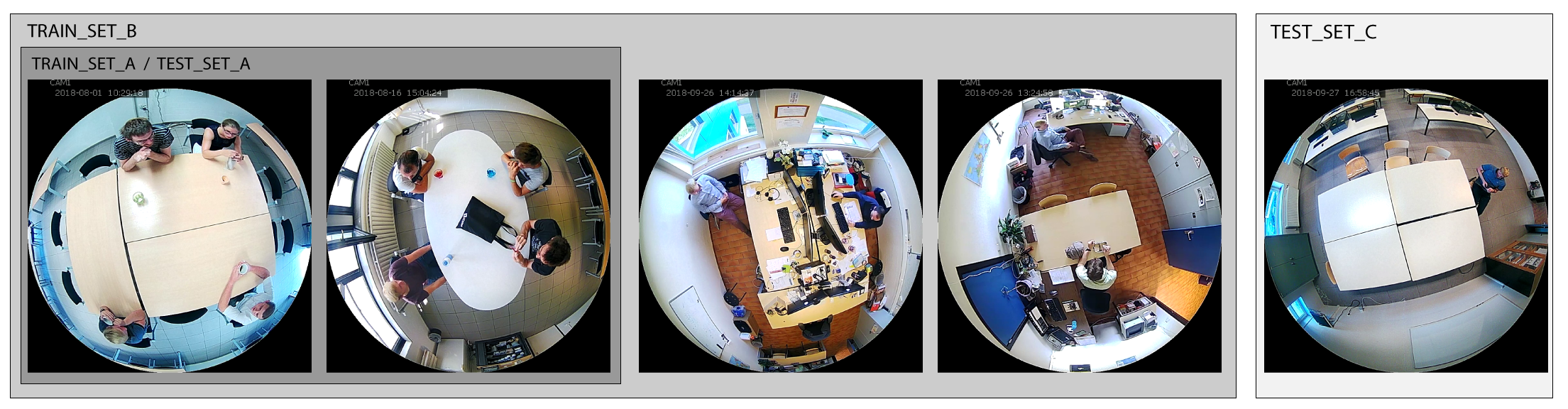}
    \caption{The left two images show the scenario used for as train set A (meeting rooms) and as test set A. The third and fourth image show two additional scenarios (flex-desks) added to train set A, to form train set B the train set B. The fifth image show an unseen scenario (meeting room) to test the generic character of the models.}
    \label{figure:dataset}
\end{figure*}

As discussed above, to maximally cover the meeting rooms with a minimum number of cameras we employ omni-directional cameras mounted at the ceiling.
To the best of our knowledge no such publicly available dataset exists which meet these criteria, we thus recorded our own dataset.
Note that high quality images are not a requirement (since we downscale the images for future processing).
We therefore recorded our videos at 15 \gls{fps} with a resolution of $876\times876$.
Our dataset consists of five different scenes (meeting rooms), as illustrated in figure \ref{figure:dataset}, and is made publicly available.
The two leftmost images show scenario A, which are divided in a separate training and a test set.
The training and test set consist of different meetings in the same room with a varying number of unique people.
As such, the scenario will be identical to the model during training and inference, but the scene is different.
The third and fourth image in figure \ref{figure:dataset} are part of the training dataset used to train model B, which is more diverse and includes a flex-desk.
Table~\ref{tbl:dataset} gives a more detailed overview of the different data set parts.

\begin{table}
\centering
\begin{tabular}{c|c|c|}
\cline{2-3}
                                              & \textbf{Images} & \textbf{Amount of people} \\ \hline
\multicolumn{1}{|c|}{\textbf{Training set A}} & 8 527        & 0-8                       \\ \hline
\multicolumn{1}{|c|}{\textbf{Training set B}} & 13 509       & 0-3                       \\ \hline
\multicolumn{1}{|c|}{\textbf{Test set A}}     & 10 100       & 0-6                       \\ \hline
\multicolumn{1}{|c|}{\textbf{Test set C}}     & 2 048        & 0-3                       \\ \hline
\end{tabular}
\caption{Details on the recorded datasets, amount of images and people per set}
\label{tbl:dataset}
\end{table}

To further evaluate the generalisability of our models we recorded test-set C consisting of a completely new scenario which was not used for training of either model.
In the next section we continue by suggesting an approach that automatically generates annotations from this dataset. 
For this we first unwarp the images before using a combination of state-of-the-art person detectors.

\section{Automatic annotations}
\label{sec:automatic_annotations}

\begin{figure*}
	\centering
	\subfloat[Unrectified image]{
	    \includegraphics[width=0.45\linewidth]{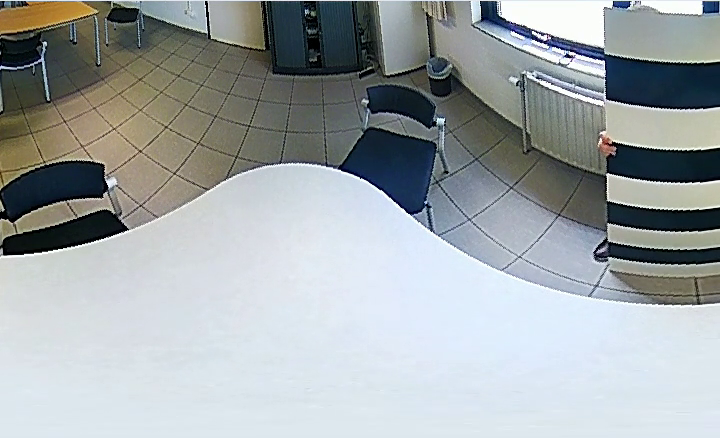}
	    \label{fig:omni_unrectified}
	}
	\qquad
	\subfloat[Rectified image]{
	    \includegraphics[width=0.45\linewidth]{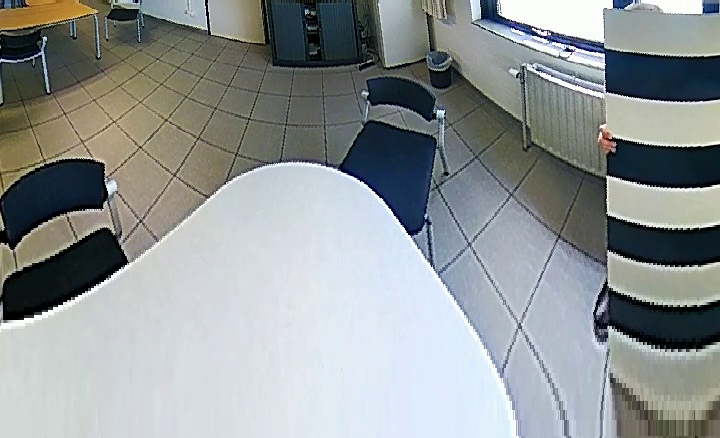}
	    \label{fig:omni_rectified}
	}
	\caption{Before and after rectifying the y-axis of the images}
\end{figure*}

\begin{figure*}
	\centering
    \includegraphics[width=\linewidth]{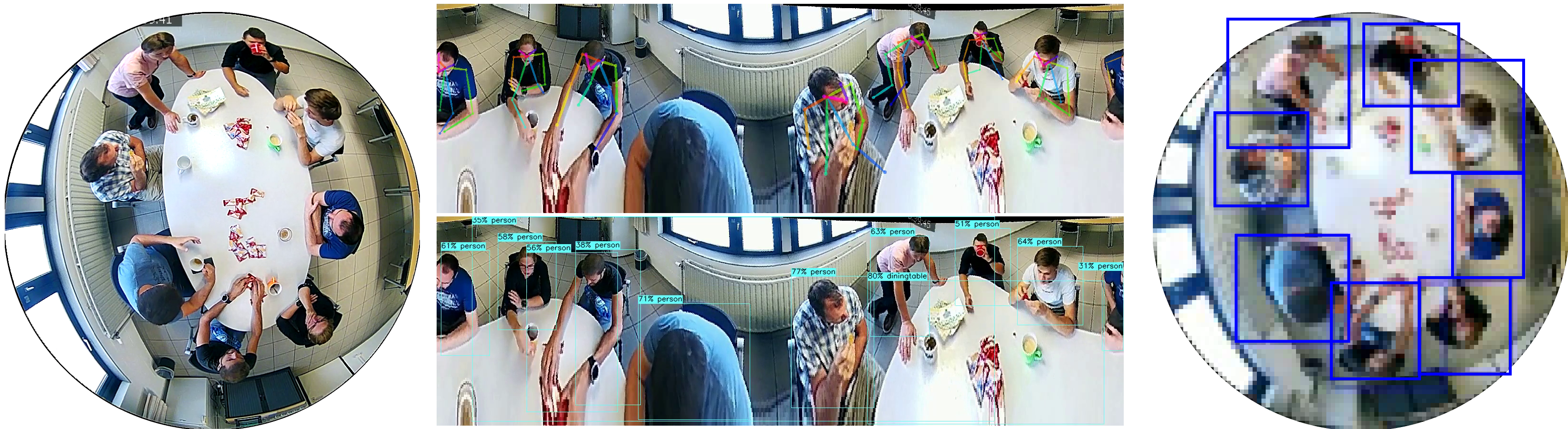}
	\caption{The omni-directional input image (left), the unwarped version with YOLOv2 detections (top centre), pose estimation detections (bottom centre) and the combined detections as automatic annotations on the low resolution image (right)}
    \label{fig:automatic_annotation_outputs}
\end{figure*}

As mentioned in section \ref{sec:dataset} we recorded our own dataset
In this section we now propose an approach that is able to automatically generate annotations based on the high-resolution input image.
This is done in a two-step process: we first unwarp/rectify the images and then perform person detection on them.
If the quality of the frame detections is insufficient, the total frame is dropped.
One quality measure we preform is looking at the amount of detections within a temporal window remaining after the confidence threshold, if the current frame has a sudden drop in detections, the frame will not be used.
This will have little to no impact, since sufficient training data is available.
%[KVB - deze details niet in de paper zetten] When we run vanilla YOLOv2 on the omni-directional image it will detect a clock, which is not unreasonable since the omni-directional camera strongly resembles a clock.

\subsection{Unwarping and rectifying}
We first start by unwarping the $876\times876$ image using the log-polar mapping method \cite{wong2011study}.
Since to the radius of the omni-directional image the unwarped image has height of 438px. As step size for the circular unwarp we chose 0.5$^\circ$, resulting in a width of 720px.

Due to the distortion of the omni-directional camera the image will be compressed along the y-axis near the outer boundary, while being stretched near the centre as illustrated in figure \ref{fig:omni_unrectified}.
In order to rectify this distortion we used a striped calibration board.
Based on these points and the desired rectified points we calculated the parameters for the following rectification equation:
\begin{equation} \label{eq:rectify_function}
    y'(y) = -0.001387 y^2 + 1.247 y - 2.007
\end{equation}

This is illustrated in figure \ref{fig:omni_rectified}.
Due to the rectification a small bottom portion of the image is lost, since this is the area with the highest distortion. Note that this is not an issue: due to our camera viewpoint the image centre coincides with the middle of the meeting tables, in which persons never need to be detected.

%This new image has an aspect ratio not frequently used in most networks, we therefore divide the image into two equal parts.
%By doing this we see that the detections improve by 15\% allowing for a stricter threshold.

\subsection{Combined detectors}

We combined both Darknet (YOLOv2 \cite{redmon2017yolo9000}) and OpenPose \cite{cao2017realtime,wei2016cpm} to detect all the people in the unwarped high-resolution images in order to generate reliable training annotations for our low-resolution detector to be trained.

As can be seen in fig. \ref{fig:automatic_annotation_outputs}, these two techniques are quite complementary to reliably detect all people in the image. We form a bounding box around the OpenPose keypoints with a margin of 25\% around each person.
In order to avoid false positives, we only accept people with more than 5 OpenPose keypoints.
When too many detections were lost after the confidence threshold we decide not to use the frame, abiding the strict policy.

Next we combine both bounding boxes using \gls{nms} \cite{neubeck2006efficient}, only leaving one detection per person.
%Afterwards we combine the detections by the boundaries by looking at the euclidean distance of the connecting corners. 
%???
Figure \ref{fig:automatic_annotation_outputs} illustrates the input omni-directional image and rectified unwarped image containing the YOLOv2 detection and the pose estimator output.
We can see that in this frame the pose estimator failed to detect the person, while YOLOv2 still found the person, showing that the combination has its benefits.
To the right we see the produced automatic annotations after downscaling the image to a low resolution version ($96\times96$).

\subsection{Validation of the automatic annotations}

\begin{figure}[ht]
    \centering
    \includegraphics[width=\linewidth]{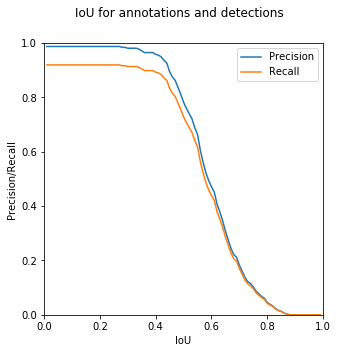}
    \caption{The precision and recall for each IoU}
    \label{figure:omni_iou}
\end{figure}

In order to validate our automatic annotation approach we manually annotated a set of 100 random images from the complete recorded dataset.
Visual analysis of the automated annotations show that there was a lot of margin due to taking the bounding box around the points that were warped back to the omni-directional frame.
The manual annotations were done properly around the persons and will have no margin.
%During automatic annotation we used a very strict policy and set several thresholds making it difficult to chose a combined confidence score as output.
%We therefore were only able to compare manual and automatic annotations by changing the minimum \gls{iou}.
Figure \ref{figure:omni_iou} shows the result of the comparison of manual and automatic annotations as the precision and recall for different values of the \gls{iou} threshold. We observe a drop occurring around an \gls{iou} of 0.4, which indeed can be explained by the different sizes of the bounding boxes. 
We therefore, for the remainder of this paper, use this \gls{iou} since we are more interested in an estimated location and not in a perfectly fitted detection.
\begin{table}
\centering
\begin{tabular}{|c|c|c|}
\cline{1-3}
\textbf{IoU}  & \textbf{0.4}    & \textbf{0.5} \\ \hline
\textbf{TP} & 299             & 244          \\ \hline
\textbf{FP} & 13              & 68           \\ \hline
\textbf{FN} & 36              & 91           \\ \hline
\textbf{P}  & 0.958           & 0.782        \\ \hline
\textbf{R}  & 0.893           & 0.728        \\ \hline
\end{tabular}
\caption{Results of the automatic analysis for IoU \{0.4; 0.5\}}
\label{tbl:omni_auto_manual_results}
\end{table}

Table \ref{tbl:omni_auto_manual_results} shows that an \gls{iou} of 0.4 misses 36 detections and introduces 13 false detections, while the \gls{iou} of 0.5 -- often used in object detection literature -- has almost 3 times as much false negatives and 5 times the amount of false positives.
The majority of these issues are only due to too large detections opposed to smaller manual annotations. Nevertheless, taken notice of their slightly less accurate positioning of bounding boxes, we conclude that our automatic annotations are reliably enough to use as training material for our low-resolution person detector network, thereby eliminating the tedious manual annotation labour work.

\section{Training a low-resolution person detector}
\label{sec:traininglowres}

To allow the resulting system to run on embedded \gls{soc} systems, we focus on low processing time from early on.
We therefore chose to perform all current experiments on YOLOv2 and kept using this network to evaluate the influence of lowering the network input resolutions.
Training a whole new network from scratch is clearly impossible given the limited amount of data.
We chose using the pretrained weights of YOLOv2 trained on the coco \cite{lin2014microsoft} dataset, that we can repurpose by using transfer learning \cite{mesnil2011unsupervised}.
Because the coco trained model is, amongst many others, trained on a class \emph{person}, we assume the network has knowledge about the visual appearance of a person, which we can inherit by fine-tuning (\emph{transfer learning}) the weights towards detecting people on low-resolution omni-directional data as well.

Traditionally, the YOLOv2 network has an input resolution between $608\times608$ and $320\times320$, yet this is not low enough for our application.
The lowest we can go in resolution with the native YOLOv2 network is $96\times96$. Indeed, the network has 5 max pooling layers with size 2 and stride 2, with 3 additional convolutions, this limits the smallest input resolution to $3\times2^5 = 96$. To investigate \emph{how low we can go}, we therefore trained between the range of 608$\ldots$96 ($448px$, $160px$ and $96px$), on which we trained two models for each input resolution (using different training data sets, to allow to investigate data set bias).

As in the original YOLOv2 implementation, during training we slightly vary the input resolution of the network around the desired input resolution such that the model becomes more scale-invariant.

For each resolution we trained two models. Model A only contains meeting room scenes, on which the camera is placed at the centre of the table.
Meanwhile, model B is trained on the same data, with addition training data (flex-desks) on which we hope to see a more generic model, since lots of additional clutter (e.g. objects on the desks) is present.
In section \ref{sec:howlowcanwego} we will discuss the results, the speed and the degree of perceived privacy of each model.

\section{Results: \emph{How low can we go?}}
\label{sec:howlowcanwego}

As explained above, for each resolution we trained two models, a first on a basic dataset A and a second on a more diverse training dataset B (as illustrated in fig. \ref{figure:dataset}). To validate these models we first used test set A (two leftmost frames in fig.~\ref{figure:dataset}), containing footage that is recorded in the same scenes as in the training set.  A second evaluation was on test set C (rightmost frame in fig.~\ref{figure:dataset}), which was recorded in a totally new room, unseen during training. We included the latter in order to test the system's generalisability towards new scenes.

\begin{figure}
	\centering
	\subfloat[Net resolution 448]{\includegraphics[width=\linewidth]{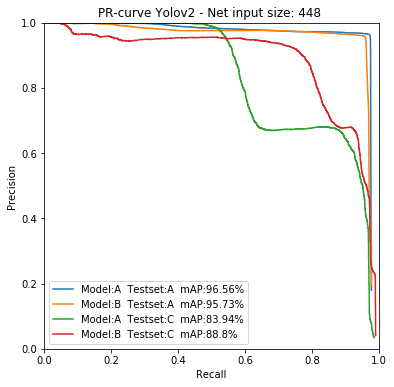}
	    \label{fig:omni_pr_curve_448}}
	    \quad
	\subfloat[Net resolution 160]{\includegraphics[width=\linewidth]{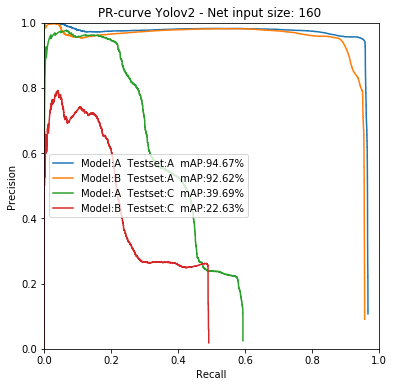}
	    \label{fig:omni_pr_curve_160}}
	    \quad
	\subfloat[Net resolution 96]{\includegraphics[width=\linewidth]{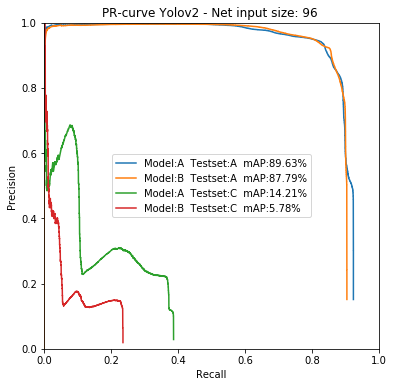}
	    \label{fig:omni_pr_curve_96}}
	\caption{PR curve of models A and B on test sets A and C with \gls{iou} 0.4}
	\label{fig:omni_pr_curves}
\end{figure}

Figure \ref{fig:omni_pr_curves} shows the precision-recall curves of these validations. We observe conclude that each model trained on training set A and validated on test set A (containing different images, but acquired in the same room as training set A) achieves high mAPs.

However, when we validate on test set C, containing an unseen scene, we notice that only the $448\times448$ model in fig.~\ref{fig:omni_pr_curve_448} performs adequately.

Remarkably, we also observe that training on a more diverse dataset (B) does not necessarily yield a more generalisable detector for unseen situations. 

Figures \ref{fig:omni_pr_curve_160} and \ref{fig:omni_pr_curve_96} illustrate that for lower input resolution the models either over-fitted on the data, or were unable to generalise to different scenarios.

Apart from accuracy we aim to determine the optimal trade-off between privacy and speed as well.
Table \ref{tbl:omni_resolution_results} shows the inference time on a V100 NVIDIA GPU, together with the amount of \gls{flops} that are needed for a single frame.
We see that the computational complexity of the $160px$ model is 8x less and that of the $96px$ model even 22x less than the original network.
This already seem reasonable when targeting an embedded system (or \gls{soc}), moreover if we lower the required \gls{fps} to one frame per minute. Indeed, for a room occupancy sensor, one measurement per minute is enough.

The rightmost column in table \ref{tbl:omni_resolution_results} shows for each resolution the equivalent blur kernel size.
Note that we made input and output videos of each model available online, together with the results of the automatic annotations\footnote{https://tinyurl.com/VISAPP2019-HLCYG}.

\begin{table}
\centering
\begin{tabular}{|c|c|c|c|}
\cline{1-4}
\textbf{Resolution}  & \textbf{\gls{fps}}    & \textbf{FLOPS}  & \textbf{Blur kernel size} \\ \hline
$448px$                  & 52                    & 34.15 Bn        & $2px$         \\ \hline
$160px$                  & 108                   & 4.36  Bn        & $5px$         \\ \hline
$96px$                   & 186                   & 1.57  Bn        & $9px$         \\ \hline
\end{tabular}
\caption{Inference speed, \gls{flops} and equivalent blur kernel size of the models on a single NVIDIA V100 GPU}
\label{tbl:omni_resolution_results}
\end{table}

\section{Conclusions}
\label{sec:conclusion}

The goal of this paper was the development of a framework which is able to detect people in a room using ceiling-mounted omni-directional cameras, allowing for occupancy optimisation in a room management system.
However, placing cameras in workplaces like meeting rooms or flex-desks is heavily regulated and most people tend to feel unease when constantly being filmed.
Therefore in this work we researched how we can use state-of-the-art detectors to detect people while ensuring their privacy.

We recorded a new publicly available dataset, containing 5 different meeting room scenes.
Because of degrading the image resolution (or using an image sensor with a very low resolution) has a positive influence on the sense of privacy, in this work we opted to develop a framework which is able to efficiently detect persons in extremely low resolution input images. The use of such low resolution images inherently ensures privacy of the individuals being recorded. We evaluated different down-scaled resolutions to determine the optimal trade-off between resolution and detection accuracy.

To avoid the need for time-consuming and expensive manual annotations we proposed an approach that is able to automatically generate new training data for the low resolution networks, based on the high resolution input images.
The validity of this approach was proven when comparing with true manual annotations.

Extensive accuracy experiments were performed. On test sets based on known scenes, the models showed an acceptable performance for all resolutions.
When tested on a similar scene with unseen data an evident declining performance with lower resolution is witnessed.
However, because of the proposed automatic annotation pipeline it remains easily possible to add additional training data for each scene. Indeed, a sensor that is newly installed in a certain room can easily acquire during the first hours some high-resolution footage, with which a room-specific low-resolution detector can quickly be (transfer) learned.  

Based upon our results we conclude that, despite the extremely low input resolution of our lowest-resolution model (96$\times$96px), our YOLOv2-based detection pipeline is still able to efficiently detect persons, even though they are not recognisable by human beings.
Our framework thus is able to serve as an efficient occupancy detection system.

Furthermore, the low input resolution allows for a lightweight network which thus is easily implementable on embedded systems while still maintaining high processing speeds.

Although the current approach is suitable to be used by the industry as is, we believe that we have not yet reached the extreme lower limit and deem it possible to decrease even further in resolution.

\section*{Acknowledgement}
\label{sec:acknowledgement}

This work is partially supported by the VLAIO via the Start to Deep Learn project.

\vfill
\newpage
\bibliographystyle{apalike}
{\small
\bibliography{main}}

\vfill
\end{document}